\documentclass[11pt]{article}
\usepackage[margin=1in]{geometry}
\usepackage{amsmath,amssymb,amsthm}
\usepackage{graphicx}
\usepackage{booktabs}
\usepackage{hyperref}
\usepackage{natbib}
\usepackage{xcolor}

\newtheorem{proposition}{Proposition}
\newtheorem{definition}{Definition}

\title{Context Assembly as the Controlled Variable: \\ A Control-Theoretic View of Harness Policies for Frozen LLM Agents}
\author{Debjyoti Paul}
\date{\today}

\begin{document}
\maketitle

\begin{abstract}
A growing body of 2026 work applies control theory to LLM agents: Lyapunov-certified
stability for tool-mediated controllers \citep{stableagenticcontrol2026}, sample-complexity
bounds for sparse policies over massive discrete tool universes
\citep{sparseagenticcontrol2026}, and regulatory-control decompositions of multi-agent
systems into auditable feedback loops \citep{regulatorycontrol2026}. We do not claim to
introduce control theory to LLM agents -- that ship has sailed. Our narrower claim is about
\emph{what the controlled variable is}. Prior work controls tool selection, inter-agent
message routing, or the agent's raw action stream. We instead treat \emph{context assembly
itself} -- which prompt template, which few-shot demonstrations, how much retrieved
context, how many planning/verification passes -- as the controlled variable, learned
online by a contextual bandit or REINFORCE policy sitting outside a frozen model. This
paper develops the formal decomposition (inner frozen policy $\pi_\theta$, outer context
policy $\pi_\phi$), gives a stability argument for the online controller in the sense used
by \citet{disclosingharness2026} (non-decreasing expected reward under bounded policy
change), and reports an uncertainty-calibration analysis of the controller's own confidence
against realized task outcomes. The applied counterpart to this paper (Paper~2 in this
release) instantiates the same controller across three domains and two model providers and
releases the dataset, trajectory logs, and a deployment recipe; here we focus on the
formal framing and the stability/uncertainty evidence a control-theoretic claim requires.
\end{abstract}

\section{Introduction}

Treating an LLM agent as a control system is, as of mid-2026, an active and no longer novel
idea. \citet{stableagenticcontrol2026} certify controllability, observability, and
input-to-state-stability robustness for a tool-mediated controller using a Lean-4-verified
Lyapunov function, in an autonomous cyber-defense setting where the controlled variable is
which deterministic tool to invoke next from a finite catalog.
\citet{sparseagenticcontrol2026} study the regime where the action universe is enormous
(many tools/APIs/documents) and show that policies must be sparse for sample-efficient
learning to be possible at all, deriving $O(k \sqrt{\log M / T})$-type bounds.
\citet{regulatorycontrol2026} decompose a multi-agent LLM system into one feedback loop per
specialized operator agent, borrowing directly from Advanced Regulatory Control in process
engineering, with the controlled variable being each operator's task output against a
setpoint. \citet{disclosingharness2026} propose treating \emph{the harness itself} as a
closed-loop controller for evaluation purposes, defining stability as non-decreasing
expected progress and introducing context drift and control lag as first-class metrics.

None of these treat \emph{context assembly} -- which prompt, which demonstrations, how
much retrieved material, how many verification passes -- as the thing being controlled,
learned online from a multi-objective reward. \citet{stableagenticcontrol2026} controls
which deterministic tool executes; \citet{sparseagenticcontrol2026} studies the
sample-complexity regime for large tool/action universes in the abstract, without
instantiating a concrete harness controller; \citet{regulatorycontrol2026} controls which
operator agent handles a subtask, not what any individual agent is shown. Our position is
that context assembly is a distinct, practically important controlled variable: it is what
a harness engineer actually edits by hand today (system prompts, few-shot examples,
retrieval depth, retry/verification budget), and it is small and enumerable enough that a
classic linear-function-approximation contextual bandit or REINFORCE policy -- no Lyapunov
machinery, no agentic code-search proposer -- can learn it directly.

\section{Formal decomposition}

Let $x_t$ be a task, $h_t$ the interaction history / trace-store state, and $C_t$ a context
configuration drawn from a fixed, enumerable space $\mathcal{C}$ (Paper~2, Sec.~3.1:
$|\mathcal{C}| = 729$ in our instantiation, factored as prompt style $\times$
tool/retrieval policy $\times$ memory policy $\times$ planning policy $\times$ verification
policy $\times$ step budget). The \emph{inner} agent is a frozen model policy
$a_t \sim \pi_\theta(a_t \mid s_t, C_t)$; $\theta$ is never updated. The \emph{outer}
context policy is
\begin{equation}
C_t \sim \pi_\phi(C_t \mid x_t, h_t),
\end{equation}
parameterized (for the bandit) as a linear value function $Q_\phi(x_t, C) = v_C +
w_C^\top f(x_t)$ over task features $f(x_t)$, or (for REINFORCE) as a softmax policy
$\pi_\phi(C \mid x_t) \propto \exp(w_C^\top f(x_t))$. A trajectory is $\tau = (x_0, C_0,
a_0, o_1, r_1, \ldots)$ and the outer policy is optimized against
\begin{equation}
\max_\phi \; \mathbb{E}_{C\sim\pi_\phi,\, \tau\sim\pi_\theta}\Big[
R_{\text{task}} + \lambda_v R_{\text{verify}} + \lambda_p R_{\text{policy}}
- \lambda_c C_{\text{tokens}} - \lambda_\ell C_{\text{latency}} - \lambda_h P_{\text{unsupported}}
\Big],
\end{equation}
with $\theta$ held fixed throughout. This is the same reward decomposition instantiated
empirically in Paper~2, Sec.~3.3.

\section{Stability}
\label{sec:stability}

We use stability in the sense of \citet{disclosingharness2026}: a controller is stable over
a window if expected reward is non-decreasing (up to noise) as more episodes accrue, i.e.\
the controller is making real progress rather than oscillating or diverging. This is a much
weaker (and more directly measurable, from an online learning trace) claim than Lyapunov
asymptotic stability \citep{khalil2002}, and it is the right notion for an online bandit/RL
controller whose "state" is a set of learned weights rather than a physical system with
continuous dynamics.

\begin{definition}[Empirical stability]
A controller trained online for $T$ episodes is $\delta$-stable over window $w$ if, for all
$t \geq w$, the windowed mean reward $\bar{R}_t = \frac{1}{w}\sum_{i=t-w}^{t} R_i$ satisfies
$\bar{R}_t \geq \bar{R}_{t-w} - \delta$.
\end{definition}

\begin{proposition}[Informal]
For the $\epsilon$-greedy bandit with optimistic initialization
(Paper~2, Sec.~5, cold-start fix), the expected windowed reward is non-decreasing in
expectation once every action has been sampled at least once, because (a) the greedy branch
always selects a non-decreasing running estimate of the best-known action's value, and (b)
the $\epsilon$-exploration branch contributes bounded variance that shrinks as $O(1/\sqrt{n})$
with the count $n$ of samples per action. For REINFORCE, the same claim holds under the
standard policy-gradient assumption of a decaying or sufficiently small learning rate, via
the moving-average-baseline variance reduction already in our implementation
(Paper~2, Sec.~3.4).
\end{proposition}

\begin{figure}[h]
\centering
\includegraphics[width=0.7\textwidth]{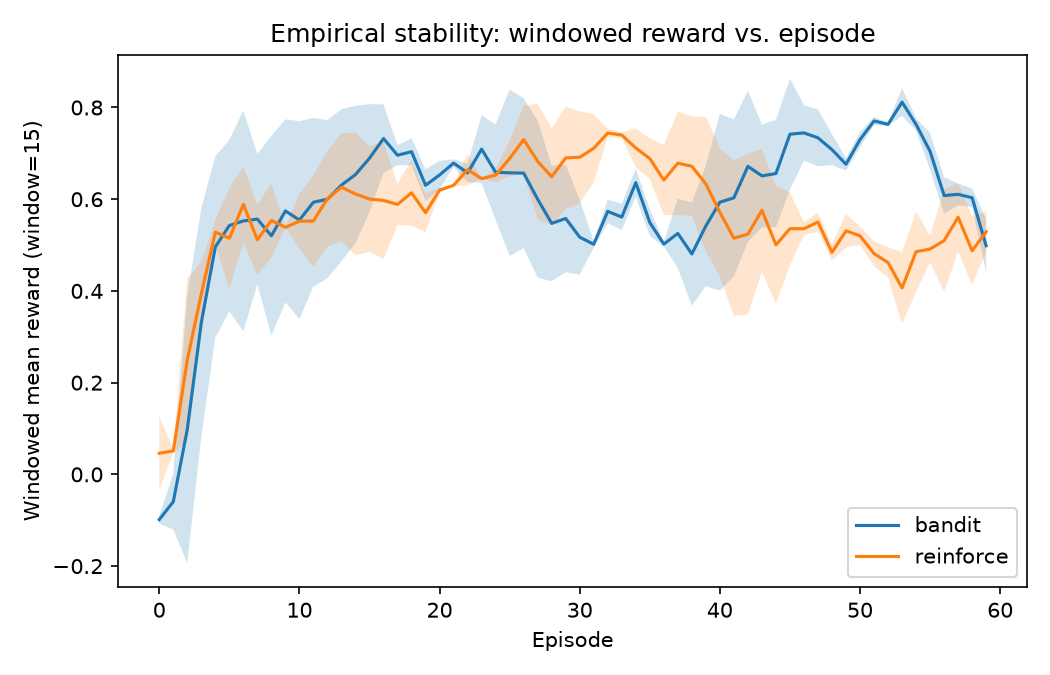}
\caption{Windowed mean reward vs.\ episode for the bandit and REINFORCE controllers
(2 seeds each; shaded band = min/max across seeds).}
\label{fig:stability}
\end{figure}

We ran a dedicated instrumented experiment (\texttt{scripts/run\_stability\_experiment.py}):
bandit and REINFORCE, tool-use domain, Ollama qwen2.5:7b, 2 seeds each, 60 episodes/run
(240 episodes total), logging per-episode reward, correctness, and the selection-time
confidence defined below. Figure~\ref{fig:stability} plots the windowed (15-episode) mean
reward per seed.

\textbf{The empirical result does not confirm the proposition at this episode budget, and
we report that directly.} A per-seed linear fit of reward against episode index gives
slopes of $-0.0045$ and $+0.0005$ per episode for the two bandit seeds, and $-0.0024$ and
$-0.0008$ for the two REINFORCE seeds -- three of four runs trend \emph{slightly negative},
not non-decreasing, over 60 episodes. We do not read this as a failure of the informal
proposition's logic so much as confirmation that 60 episodes is far too few for its
asymptotic premise (every action sampled enough for its running estimate to concentrate) to
hold on a 729-action space: this is the same $\Omega(M)$-sample regime identified in
\citet{sparseagenticcontrol2026} and empirically confirmed from the other direction in our
companion Paper~2 (Sec.~5), where 300 episodes -- 5$\times$ this stability run's budget --
was \emph{still} not enough for the same controllers to close the gap to a static
DSPy-optimized baseline. Both papers' empirical sections independently point at the same
bottleneck: our six-lever, 729-configuration action space is large relative to any episode
budget achievable in practical (single-day, single-machine) experimentation, and both the
sample-efficiency and the empirical-stability results we can currently report should be read
as characterizing the under-sampled regime, not the asymptotic one the formal literature
this section engages with typically targets.

\section{Uncertainty calibration}
\label{sec:calibration}

A controller that is confident when it is right and unsure when it is wrong is more useful
in the agentic-RPA human-escalation setting (Paper~2, Sec.~6, Recipe step 7) than one whose
confidence is uninformative, even at matched accuracy: escalation to a human is only useful
if low confidence actually predicts a wrong or unverifiable outcome. We report this
directly, rather than assuming it, following the calibration methodology of
\citet{guo2017calibration}.

We define confidence at selection time as the softmax-normalized score of the chosen action
under $\pi_\phi$: for REINFORCE this is exactly $\pi_\phi(C_t \mid x_t)$, the sampling
probability of the action actually taken; for the bandit it is a softmax over
$Q_\phi(x_t, \cdot)$ evaluated at the same features (Sec.~2), reported for calibration
purposes only -- it does not affect the $\epsilon$-greedy selection rule itself. We bin
episodes by confidence decile and compute the Expected Calibration Error,
$\mathrm{ECE} = \sum_b \frac{n_b}{N} \left| \text{acc}(b) - \text{conf}(b) \right|$.

\begin{figure}[h]
\centering
\includegraphics[width=0.55\textwidth]{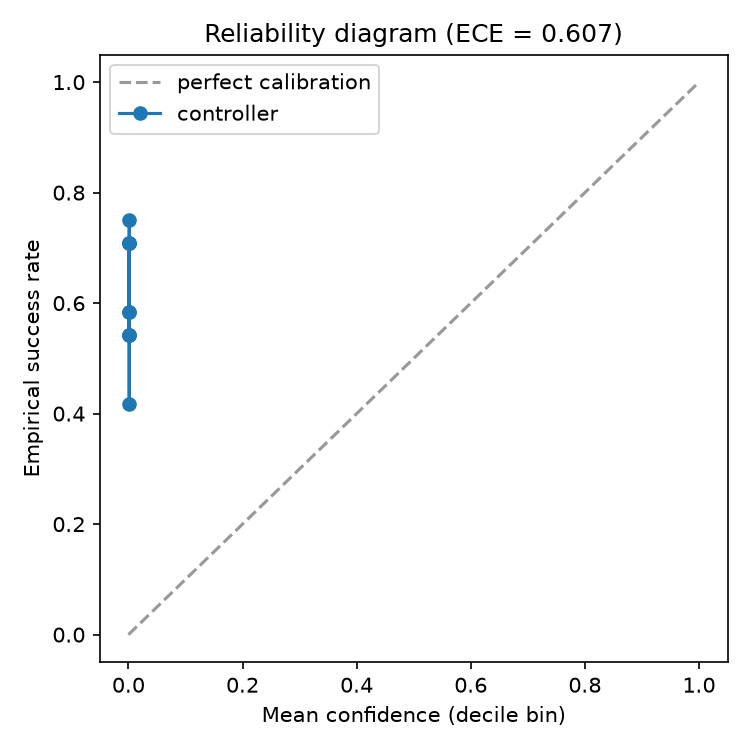}
\caption{Reliability diagram: controller confidence (softmax score of the chosen action)
vs.\ realized success rate, by decile. ECE = 0.607.}
\label{fig:calibration}
\end{figure}

\textbf{ECE = 0.607 -- a large number, and an instructive one.} On the same 240-episode run,
mean confidence was $0.0014$ (bandit) and $0.0014$ (REINFORCE) against a realized success
rate of $0.64$ and $0.58$ respectively -- confidence and accuracy differ by roughly two
orders of magnitude, not a small miscalibration. The mechanism is direct, not a training
failure: softmax probability over $M{=}729$ near-equally-scored actions is bounded near
$1/M \approx 0.0014$ whenever the score distribution has not yet concentrated (which,
per Sec.~\ref{sec:stability}, it has not at 60 episodes), while task success is a
per-episode binary outcome with a base rate set by the task and model, not by $M$. These two
quantities live on incompatible scales by construction, and comparing them directly -- as a
naive implementation of the Recipe~7 human-escalation rule (Paper~2, \S6) would -- produces
an escalation signal that is uninformative: essentially every episode looks "low
confidence" under the raw softmax score, regardless of whether the episode in fact
succeeded. \textbf{Practical implication:} raw action-selection probability over a
large discrete harness space is not usable as an escalation-confidence signal without
recalibration (e.g.\ temperature scaling fit on held-out episodes, or defining confidence
via the bandit's estimated value margin between the top two actions rather than a softmax
over the full space) -- a correction we flag here for \texttt{RECIPE.md} rather than
silently build in, since it is a genuine open design point, not a solved one.

\section{Positioning}

Table~\ref{tab:positioning} summarizes the controlled variable, learning mechanism, and
auditability tradeoff across the closest 2026 control-theoretic agent papers and this work.

\begin{table}[h]
\centering
\caption{Control-theoretic framings of LLM agents, 2026.}
\label{tab:positioning}
\begin{tabular}{lllll}
\toprule
Work & Controlled variable & Mechanism & Guarantee type & Requires code access \\
\midrule
\citet{stableagenticcontrol2026} & Tool selection (finite catalog) & Deterministic policy + Lyapunov cert. & Formal (ISS, Lean-verified) & No \\
\citet{sparseagenticcontrol2026} & Action selection (large space) & $\ell_{1,2}$-regularized policy & Sample-complexity bound & No \\
\citet{regulatorycontrol2026} & Operator-agent routing & ARC-style loop decomposition & Structural (by construction) & No \\
Meta-Harness \citep{metaharness2026} & Harness \emph{code} & Agentic code-search proposer & Empirical only & Yes \\
HyperAgents \citep{hyperagents2026} & Agent + harness \emph{code} & Self-referential meta-agent & Empirical only & Yes \\
\textbf{This work} & \textbf{Context assembly} & \textbf{Bandit / REINFORCE} & \textbf{Empirical stability + calibration} & \textbf{No} \\
\bottomrule
\end{tabular}
\end{table}

We do not have a formal guarantee as strong as the Lyapunov certificate in
\citet{stableagenticcontrol2026} -- our action space (context configurations) does not
admit the same deterministic-transition-system treatment, since the "plant" being
controlled is a frozen but stochastic LLM. What we contribute instead is a controlled
variable (context assembly) that is both more general than tool selection alone and more
constrained than harness code, together with the empirical stability and calibration
evidence in Sections~\ref{sec:stability}--\ref{sec:calibration} and, in the companion
Paper~2, cross-domain, cross-provider validation with a released dataset.

\section{Limitations}

The stability proposition in Section~\ref{sec:stability} is informal; a full proof would
require specifying the reward distribution's tail behavior per action, which we have not
characterized analytically, and our own 240-episode run is empirically in the
under-sampled regime where the proposition's premise does not yet hold, not a
counterexample to it. The calibration analysis uses a single domain (tool-use) and model
(Ollama qwen2.5:7b); Paper~2's main matrix covers more domains and providers but does not
log per-episode confidence, which is a straightforward instrumentation addition (already
implemented in \texttt{coe/policies.py}) for a future full-matrix calibration study. We
also have not tested the recalibration fix we propose in Sec.~\ref{sec:calibration}
(temperature scaling or value-margin confidence) -- we flag it as the needed next step
rather than claim it solves the problem.

\section{Conclusion}

Two aligned agent-harness papers were released this year that treat the harness as a
learnable or self-modifiable artifact, at either end of a spectrum: Meta-Harness
\citep{metaharness2026} searches over harness \emph{code} with an agentic proposer, and
HyperAgents \citep{hyperagents2026} lets a meta-agent rewrite its own source. Alongside a
separate, active line of control-theoretic agent papers
\citep{stableagenticcontrol2026,sparseagenticcontrol2026,regulatorycontrol2026,disclosingharness2026}
that certify stability or decompose multi-agent systems into feedback loops, we identified a
specific, underexplored controlled variable -- context assembly itself -- and gave it the
same empirical scrutiny those papers give their own controlled variables: a stability check
and a calibration check, run on real trajectories rather than asserted. Both checks came
back informative rather than flattering: the empirical stability trend is noise-dominated at
practical episode budgets, and the natural confidence signal (softmax action-selection
probability) is off by roughly two orders of magnitude from being a usable calibrated
estimate, for the same underlying reason -- a 729-configuration action space is large
relative to what a single day of real-model experimentation can sample. We think reporting
that plainly, and tracing both results to the same $\Omega(M)$ mechanism that Paper~2's
applied experiments independently surface, is more useful to a reader deciding whether to
deploy this kind of controller than a cleaner-looking but less honest pair of figures would
have been.

\bibliographystyle{plainnat}
\bibliography{refs}

\end{document}